\documentclass[sigconf]{acmart}
\renewcommand\footnotetextcopyrightpermission[1]{} 

\setcopyright{none}
\renewcommand\footnotetextcopyrightpermission[1]{}
\acmConference[]{}{}{}
\AtBeginDocument{%
  \fancyhead[R]{}%
}

\usepackage{amsmath}
\usepackage{amsfonts}
\usepackage{algorithmic}
\usepackage{graphicx}
\usepackage{textcomp}
\usepackage{xcolor}
\usepackage{algorithm}
\usepackage{booktabs}
\usepackage{xcolor}

\newcommand\flavio[1]{}
\newcommand\nilesh[1]{}
\newcommand\onat[1]{}
\newcommand\jangseon[1]{}
\newcommand\team[1]{}
\newcommand\Q[1]{}
\newcommand\done[1]{}
\usepackage{multirow}
\usepackage{url}
\begin{document}

\title{QMC: Efficient SLM Edge Inference via Outlier-Aware Quantization and Emergent Memories Co-Design}

\author{Nilesh Prasad Pandey}
\authornote{Both authors contributed equally to this research.}
\email{nppandey@ucsd.edu}
\affiliation{%
  \institution{University of California San Diego}
  \city{San Diego}
  \state{CA}
  \country{USA}
}

\author{Jangseon Park}
\email{jap036@ucsd.edu}
\authornotemark[1]
\affiliation{%
  \institution{University of California San Diego}
  \city{San Diego}
  \state{CA}
  \country{USA}
}

\author{Onat Gungor}
\email{ogungor@ucsd.edu}
\affiliation{%
  \institution{University of California San Diego}
  \city{San Diego}
  \state{CA}
  \country{USA}
}

\author{Flavio Ponzina}
\email{fponzina@sdsu.edu}
\affiliation{%
  \institution{San Diego State University}
  \city{San Diego}
  \state{CA}
  \country{USA}
}

\author{Tajana Rosing}
\email{tajana@ucsd.edu}
\affiliation{%
  \institution{University of California San Diego}
  \city{San Diego}
  \state{CA}
  \country{USA}
}









\begin{abstract}
  Deploying Small Language Models (SLMs) on edge platforms is critical for real-time, privacy-sensitive generative AI, yet constrained by memory, latency, and energy budgets. Quantization reduces model size and cost but suffers from device noise in emerging non-volatile memories, while conventional memory hierarchies further limit efficiency. SRAM provides fast access but has low density, DRAM must simultaneously accommodate static weights and dynamic KV caches, which creates bandwidth contention, and Flash, although dense, is primarily used for initialization and remains inactive during inference. These limitations highlight the need for hybrid memory organizations tailored to LLM inference. We propose \textbf{Outlier-aware Quantization with Memory Co-design (QMC)}, a retraining-free quantization with a novel heterogeneous memory architecture. QMC identifies inlier and outlier weights in SLMs, storing inlier weights in compact multi-level Resistive-RAM (ReRAM) while preserving critical outliers in high-precision on-chip Magnetoresistive-RAM (MRAM), mitigating noise-induced degradation. On language modeling and reasoning benchmarks, QMC outperforms and matches state-of-the-art quantization methods using advanced algorithms and hybrid data formats, while achieving greater compression under both algorithm-only evaluation and realistic deployment settings. Specifically, compared against SoTA quantization methods on the latest edge AI platform, QMC reduces memory usage by $6.3\times$--$7.3\times$, external data transfers by $7.6\times$, energy by $11.7\times$, and latency by $12.5\times$ when compared to FP16, establishing QMC as a scalable, deployment-ready co-design for efficient on-device inference.
\end{abstract}



\keywords{Small Language Models (SLMs), Hardware-Algorithm Co-design, Emerging Memories, Quantization}


\maketitle

\section{Introduction}

Large Language Models (LLMs) have significantly advanced generative AI, enabling strong performance across diverse language tasks~\cite{team2025gemma,touvron2023llama,zhang2022opt}. However, their scale, often billions of parameters, poses major challenges for deployment on resource-constrained edge devices. To address this, the community has shifted attention toward Small Language Models (SLMs) and hybrid architectures under 3B parameters, e.g., LLaMA (3B)~\cite{touvron2023llama}, Phi-2 (1.5B)~\cite{javaheripi2023phi}, Mamba (350M - 2.8B)~\cite{gu2023mamba}, and Hymba (1.5B)~\cite{dong2024hymba}). SLMs strike a practical balance between accuracy and the tight memory, bandwidth, and power budgets of the edge platforms, where larger LLMs models are infeasible to deploy. In this work, we therefore focus on these class of models as our primary target for on-device edge inference.

Although SLMs reduce parameter counts, deploying them on edge platforms such as the NVIDIA Jetson AGX Orin~\cite{orin_agx_module_ds} remains difficult due to memory bottlenecks. SRAM is fast but scarce, DRAM must serve both weights and KV caches causing bandwidth contention, and Flash is dense but inactive during inference. These constraints motivate a hardware–algorithm co-design approach to better align model behavior with the underlying memory hierarchy.

Quantization~\cite{nagel2021white,lin2024awq,pakrishnamoorthi2018quantizing,liu2024spinquant,pandey2023practical,pandey2023softmax} is a key algorithmic technique that compresses model parameters into low-precision formats. In particular, 4-bit integer (INT4)~\cite{lin2024awq,frantar2022gptq,nagel2020up} quantization is widely used for edge deployment due to its reduced memory usage compared to Floating Point 16 (FP16). However, existing approaches face the following challenges: Quantization-Aware Training (QAT)~\cite{liu2023llm,chen2024efficientqat,bhardwaj2024oh} requires costly retraining and access to proprietary data, while Post-Training Quantization (PTQ)~\cite{lin2024awq,frantar2022gptq} often suffers from accuracy degradation, as it relies on calibration datasets whose quality and availability strongly affect performance, and shows poor adaptability to emerging architectures such as state-space or hybrid models~\cite{chiang2024quamba}, limiting its applicability across LLM variants.

Beyond algorithmic constraints, edge deployment faces fundamental memory hierarchy limitations. Traditional architectures face significant performance bottlenecks in LLM inference due to their reliance on DRAM for storing both static model weights and dynamic activations. This memory hierarchy creates substantial bandwidth limitations under the read-intensive workloads that characterize LLM inference~\cite{lin2024awq}. During typical decoding operations, DRAM read activity remains consistently high throughout each step, while compute utilization falls below 20\% as processing units wait for memory transfers~\cite{recasens2025mind}. Although quantization reduces the memory footprint, existing post-training methods assume ideal hardware, which leaves them both sensitive to calibration procedures and prone to degradation when exposed with device-level variability~\cite{frantar2022gptq,lin2024awq} during deployment.

\begin{figure}[t]
  \centering
  \includegraphics[width=0.75\columnwidth]{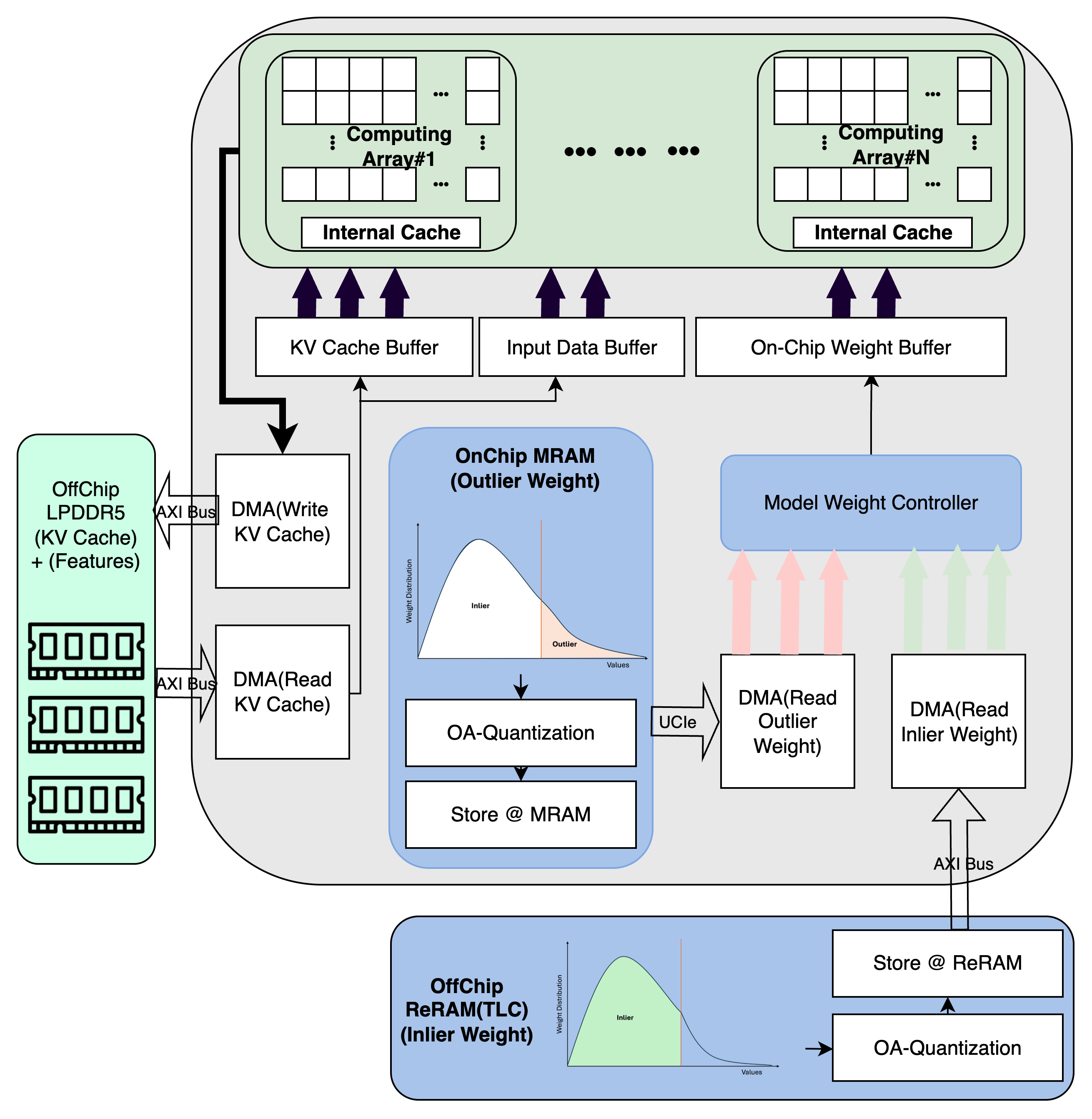}
        \caption{\textbf{QMC Architecture:} Heterogeneous memory system with on-chip MRAM for outlier weights, off-chip MLC ReRAM for inlier weights, and LPDDR5 for KV caches, coordinated by a unified Model Weight Controller.}
  \label{fig:architecture}
  \vspace{-1.5\baselineskip} 
\end{figure}

In this work (Figure ~\ref{fig:architecture}), we propose augmenting the conventional memory hierarchy with emerging non-volatile memories (NVMs) to directly store and serve model parameters. Our hardware-software co-design leverages on-chip MRAM for high-magnitude outlier weights, providing enhanced capacity and speed compared to SRAM-only solutions while ensuring reliability for sensitive parameters. The bulk of quantized inlier weights are mapped to dense multi-level cell (MLC) ReRAM, exploiting its superior density~\cite{song2025hybrid, Seitzer2024FlashForward}. DRAM is reserved exclusively for dynamic data such as activations and KV caches. Building on this, we introduce Outlier-Aware Quantization with Memory Co-design (QMC), a data-free algorithm that partitions weights into inliers and outliers, applies noise-robust quantization to inliers stored in ReRAM, and preserves outliers in MRAM, jointly improving efficiency and ensuring robustness against inherent device-level noise. This design offers three advantages: (i) parallel bandwidth through concurrent DRAM and NVM access; (ii) reduced DRAM capacity and refresh power; and (iii) elimination of Flash, since NVMs retain weights across power cycles. Our key contributions are as follows: 
\begin{itemize}
    \item We propose \emph{QMC}, a post-training quantization (PTQ) method that preserves high-magnitude outlier weights while aggressively compressing inliers. QMC outperforms or at par with state-of-the-art weight-only baselines~\cite{lin2024awq,frantar2022gptq} and, unlike prior work, explicitly maintains accuracy under device-level noise and intrinsic ReRAM variability.

    \item Our framework is both general and practical, accommodating small language and hybrid models without the need for model-specific software adaptations. In contrast to existing methods~\cite{lin2024awq,frantar2022gptq} that encounter compatibility issues~\cite{mit-han-lab_llm-awq_issue285}, it allows straightforward, out-of-the-box deployment.

    \item Lastly, we introduce a \emph{heterogeneous memory co-design} that maps high-precision outlier weights to on-chip MRAM and aggressively compressed inliers to dense multi-level ReRAM. This co-design delivers substantial system-level gains, achieving up to $11\times$ energy reduction, $13\times$ latency improvement, and $7\times$ memory savings compared to state-of-the-art quantization methods evaluated on the latest edge AI platforms~\cite{orin_agx_module_ds}. Furthermore, when compared to the state-of-the-art NVM co-design method, eMEMs~\cite{mukherjee2021case}, QMC reduces energy consumption by $1.35\times$, latency by $1.9\times$, and memory usage by $1.82\times$.
\end{itemize}





\section{Background and Related Work}
Recent advances in small language models (SLMs) have been enabled by state-space architectures such as Mamba~\cite{gu2023mamba} and hybrid variants like Hymba~\cite{dong2024hymba}, which replace quadratic attention with compact recurrent states and linear-time operations. While pure State Space Models (SSMs)~\cite{gu2023mamba} generally underperform transformers on complex reasoning tasks, hybrid designs~\cite{dong2024hymba} that combine the efficiency of SSMs with the accuracy of selective attention offer a promising balance of accuracy and efficiency for edge deployment. SLMs evolve quickly, and the rise of architecture-specific accelerators~\cite{li2024marca} makes portability without losing efficiency important.


Quantization has become a central technique for enabling inference under strict edge resource constraints. By mapping floating-point weights to low-bit integer representations, quantization reduces memory footprint and bandwidth, but at the cost of rounding and clipping errors. These errors are particularly amplified in large models due to heavy-tailed weight distributions with significant 
outliers~\cite{dettmers2022gpt3,zhao2019improving}. Quantization-aware training (QAT) can mitigate these issues, but requires access to training data and substantial compute resources. In contrast, post-training quantization (PTQ) methods such as GPTQ~\cite{frantar2022gptq} and AWQ~\cite{lin2024awq} avoid retraining, yet they remain highly sensitive to calibration data. Furthermore, PTQ methods typically assume stable memory systems, but in practice, heterogeneous non-volatile memories (NVMs) introduce device-level noise and variability that amplify quantization error and cause further degradation. These limitations are particularly problematic for SLMs, where sequential recurrence amplifies error accumulation~\cite{pierro2024mamba}. In parallel, recent works have also explored hybrid numerical formats such as MXINT4~\cite{sharify2024post}, which are orthogonal to quantization algorithms. Rather than introducing new quantization schemes, these formats exploit mixed bases to achieve better accuracy–compression trade-offs, complementing existing techniques.

Complementary to algorithmic compression, hardware–software co-design with emerging NVMs offers an attractive path toward efficient and robust deployment. Technologies such as STT-MRAM, SOT-MRAM, and ReRAM provide compatibility with advanced nodes, low-voltage operation, and inherent non-volatility~\cite{yu2024semiconductor,legtchenko2025managed,Jiang2025Emerging}. MRAM offers high reliability and noise immunity for accuracy-critical parameters~\cite{wolters2024memoryalluneed}, while ReRAM delivers superior density through multi-level cell storage for noise-tolerant weights~\cite{song2025hybrid}. Both technologies are approaching commercial maturity, with embedded solutions demonstrated at 16–22nm nodes~\cite{you2024vlsi}, whereas Phase-Change Memory (PCM) and Ferroelectric RAM (FeRAM) continue to face scaling and endurance limitations~\cite{huang2024progress}. However, prior co-design efforts utilizing emerging NVMs, such as eMEMs~\cite{mukherjee2021case}, primarily explored homogeneous memory system architectures solely for storing model weights. Another direction of research has focused on NVM-based acceleration for neural workloads~\cite{song2025hybrid,jang2024big,song2025stimc}, yet the integration of heterogeneous NVMs with quantized SLMs remains largely unexplored, presenting opportunities for efficient inference without architectural redesign.

\section{QMC: Outlier-Aware Quantization with Memory Co-design}
\subsection{Overview}
The proposed QMC constitutes a co-design framework that integrates outlier-aware quantization with a hybrid memory organization. Unlike approaches that rely on specialized hardware, QMC can be seamlessly deployed on general-purpose accelerators, such as GPUs, as well as compute units commonly used for SSM acceleration, because it only changes how weights are quantized and stored and not how they are executed. This generality eliminates the necessity for custom hardware designs while ensuring scalability and efficiency. Figure~\ref{fig:architecture} illustrates the QMC architecture, which integrates an outlier-aware quantization algorithm with a heterogeneous memory system. This hierarchy employs on-chip MRAM for reliability-critical weights requiring high precision, off-chip MLC ReRAM for aggressively quantized, noise-tolerant weights, and LPDDR5 for KV caches, all coordinated by a unified Model Weight Controller. The subsequent sections detail the criteria for weight classification and the specific quantization algorithm employed.

\subsection{Outlier-Aware Quantization Methodology}



Weight distributions in LLMs exhibit significant heterogeneity, with a small fraction of large-magnitude weights contributing disproportionately to model accuracy~\cite{yu2024super,zhao2019improving}. Prior PTQ studies observe that layers are particularly sensitive to perturbations on these large-magnitude weights, and that they contribute most of the overall quantization error~\cite{nagel2021white,frantar2022gptq,dettmers2022gpt3}. This makes weight magnitude a simple, data-free importance metric by identifying and protecting the outlier weights, preserving quantization resolution where it has the greatest impact, without requiring calibration data. We exploit this property through a dual-precision quantization strategy that preserves these critical outliers in higher precision while aggressively compressing the remaining inlier weights.

\textbf{Weight Partitioning.} We partition each weight tensor $W$ into two disjoint sets based on magnitude thresholding:
\begin{equation}
    W_{\text{out}} = \{w \in W : |w| > \tau\}, \quad W_{\text{in}} = W \setminus W_{\text{out}},
\end{equation}
where the threshold $\tau$ is chosen such that $|W_{\text{out}}| = \rho |W|$ for a fixed outlier ratio $\rho$. We then apply this same global ratio to every layer, selecting the top-$\rho$ fraction of weights by magnitude. This simple, uniform rule is both efficient and consistently yields strong accuracy and energy benefits (refer Sections ~\ref{subsec:algo_only} and ~\ref{subsec:sys_only}), making more complex layer-wise strategies unnecessary.

\textbf{Heterogeneous Quantization Strategy.} The partitioned weights undergo different quantization processes:
\begin{equation}
Q(W) =
\begin{cases}
Q_{\text{high}}(W_{\text{out}}), & W \in W_{\text{out}}, \\
Q_{\text{low}}(W_{\text{in}}), & W \in W_{\text{in}}.
\end{cases}
\end{equation}


The outlier quantizer $Q_{\text{high}}$ retains higher precision for accuracy-critical weights, while $Q_{\text{low}}$ aggressively compresses inliers. This dual-precision design naturally maps outliers to reliable, low-latency MRAM and inliers to dense ReRAM. The specifics of device choices, bit-widths, and integration are discussed in next subsection.

\subsection{QMC System Architecture}
\label{subsec:memory-arch}
\subsubsection{System Overview}
Our QMC framework employs a heterogeneous memory architecture optimized for the static nature of model weights during inference. Unlike activations or KV caches, which grow dynamically and require frequent updates in DRAM, model weights are static during inference. This observation motivates the use of non-volatile memories (NVMs) for weight storage. By exploiting their high density, low read latency, and low power consumption, we improve energy efficiency and performance over DRAM-only systems while avoiding endurance and write latency issues, since writes occur only during infrequent model updates.

\subsubsection{Heterogeneous Memory Architecture}


QMC heterogeneous memory hierarchy, which optimizes performance, area, and power by distributing resources across on-chip and off-chip domains under a unified Model Weight Controller. Embedded MRAM, integrated via 2.5D chiplet technology and accessed through UCIe~3.0 (64GT/s per IO, 64 IOs), stores reliability-critical outlier weights to ensure low latency and robustness. Conversely, aggressively quantized inlier weights are allocated to off-chip MLC ReRAM to leverage its high density. This off-chip placement is necessary as the target model requires approximately 100.65~mm$^2$ in 3-bit MLC mode. The ReRAM module interfaces via a high-speed memory bus (3.3GHz, 64-byte IO), while LPDDR5 serves KV caches and activations.

\subsubsection{Performance and Power Optimization}
We determine the memory allocation using a heuristic approach guided by two constraints. First, weight loading latency is modeled as
\begin{equation}
T = t_{\text{access}} + \frac{s}{b} + t_{\text{queue}}, \quad 
T_{\text{final}} = \max(T_{\text{mram}}, T_{\text{reram}}) +T_{\text{sync}},
\end{equation}
where $t_{\text{access}}$ denotes the intrinsic device access latency, $s$ is the data size to be transferred, $b$ represents the effective bandwidth, and $t_{\text{queue}}$ accounts for queuing delays. Since MRAM and ReRAM weights are fetched concurrently and merged before being delivered to the compute node, the overall latency is dominated by the slower of the two devices. The term $T_{\text{sync}}$ captures the synchronization overhead arising from clock domain crossing via dual-clock buffers, which introduces 2-4 clock cycles of latency ~\cite{apperson2007scalable}. Given the large-scale transfers of LLM workloads, bandwidth remains the dominant contributor to the effective latency.
Second, the power budget constraint is enforced as
\begin{equation}
\begin{split}
P_{\text{budget}} > & \; BW_{\text{mram}} \times (E_{\text{read,mram}} + E_{\text{network}}) \\
& + BW_{\text{reram}} \times (E_{\text{read,reram}} + E_{\text{network}}),
\end{split}
\end{equation}
where $P_{\text{budget}}$ is the total available memory power budget, $BW_{\text{mram}}$ and $BW_{\text{reram}}$ denote the sustained bandwidths of MRAM and ReRAM, respectively, $E_{\text{read,mram}}$ and $E_{\text{read,reram}}$ are the per-bit read energies, and $E_{\text{network}}$ is the per-bit energy overhead of the interconnect.
We perform a design space exploration across discrete MRAM–ReRAM bandwidth configurations, filtering out those that violate either constraint, and select the configuration that minimizes inference latency while staying within the power budget.

\begin{figure}
  \centering
  \includegraphics[width=0.9\columnwidth]{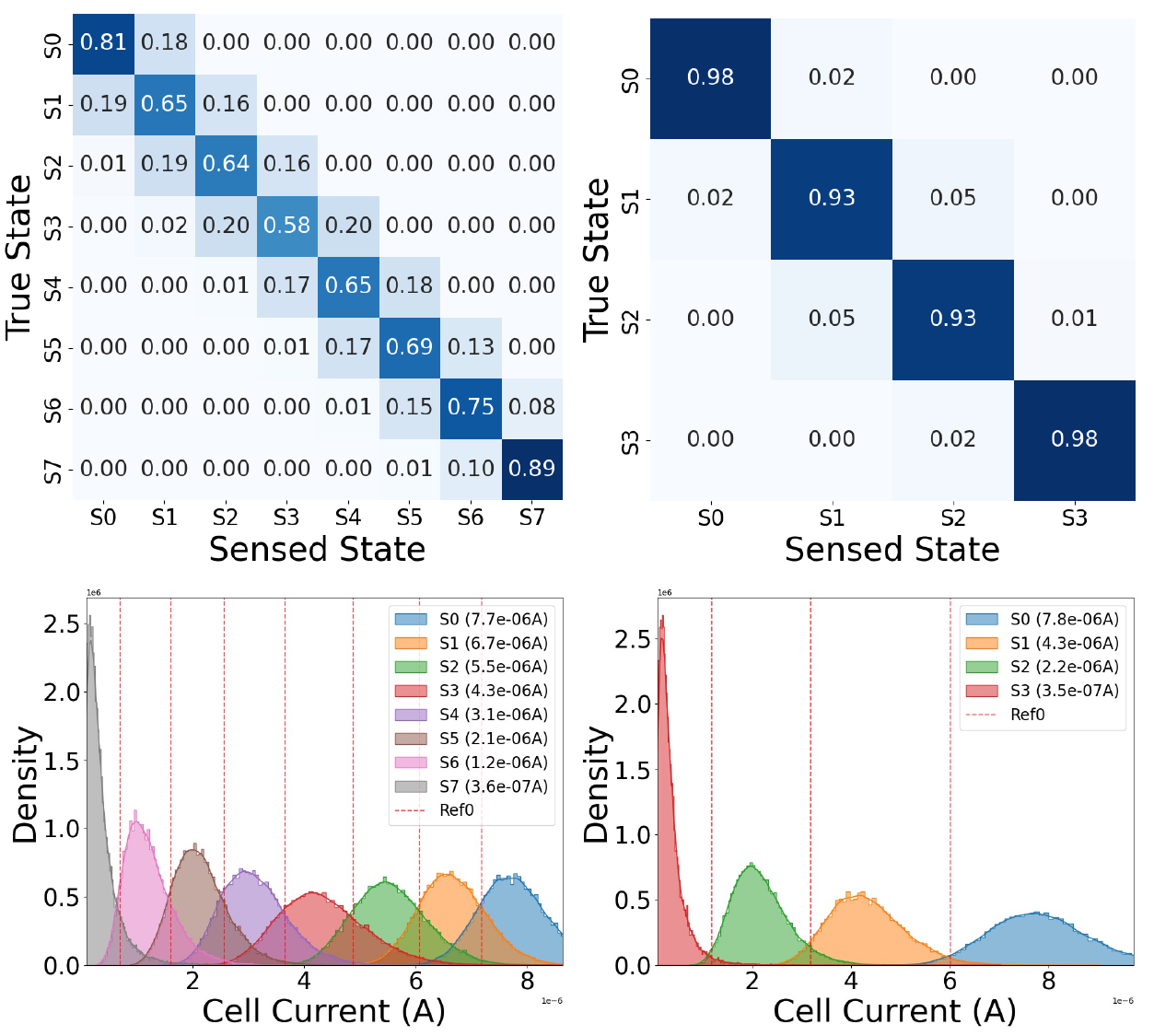}
    \caption{\textbf{MLC ReRAM Error Analysis.} Confusion matrices and read current distributions for 3-bit (left, states S0-S7) and 2-bit (right, states S0-S3) MLC ReRAM, showing state detection accuracy and current separation between levels.}
  \label{fig:error_model}
  \vspace{-0.5cm} 
\end{figure}

\subsection{Robust Quantization under ReRAM Noise}

While the heterogeneous memory architecture addresses the storage and performance requirements of outlier-aware quantization, the use of MLC ReRAM for inlier weights introduces additional challenges. Unlike conventional digital memories, MLC ReRAM exhibits unstable conductance states that can amplify quantization error. To model this effect, we utilized parameters from fabricated 40nm MLC ReRAM~\cite{fan2024efficient}. Figure~\ref{fig:error_model} presents the current distributions and confusion matrices for 2-bit and 3-bit MLC, showing the trade-off between storage density and sensing accuracy. We model this variability as discrete perturbations on quantized weights, where $e \in \{-\Delta (s),0,+\Delta(s)\}$ with probabilities $(p_{-},p_{0},p_{+})$ determined by the device BER, and $\Delta (s)$ denotes the quantization step. 
\paragraph{Noise-aware scale optimization:} As shown in Algorithm~\ref{alg:QMC}, for each inlier tensor $W_{\text{in}}$, the scale $s_{\text{ReRAM}}$ jointly controls the quantizer $Q(\cdot; s)$ and the step size $\Delta(s)$. Using the discrete perturbation model above, each quantized weight experiences an additive noise $e \in \{-\Delta(s), 0, +\Delta(s)\}$ derived from the measured BER of the MLC ReRAM device~\cite{fan2024efficient}. Hence, the expected distortion in Step~2 is

\begin{algorithm}[t]
\caption{Outlier-Aware Robust Quantization}
\label{alg:QMC}
\begin{algorithmic}[1]
\REQUIRE Weight $W$, outlier ratio $\rho$, bitwidths $(b_{\text{ReRAM}}, b_{\text{MRAM}})$
\ENSURE Quantized weights $\tilde{W}$

\STATE \textbf{Step 1: Outlier Selection} 
Set threshold $\tau$ as top $\rho\%$ of $|W|$, then partition  
$W_{\text{out}}=\{w \in W:|w|>\tau\}$ and $W_{\text{in}}=W\setminus W_{\text{out}}$.

\STATE \textbf{Step 2: Quantized Inliers stored in ReRAM} Optimize scale $s_{\text{ReRAM}}$ by solving:
\[
s_{\text{ReRAM}}^* = \arg\min_{s}\;\mathbb{E}_e\!\left[\|W_{\text{in}}-(Q(W_{\text{in}};s)+e)\|^2\right],
\]
where $e$ is device noise. Quantize inliers as
\[
W^*_{\text{in}}=Q(W_{\text{in}};s_{\text{ReRAM}}^*).
\]

\STATE \textbf{Step 3: Quantization of Outliers stored in MRAM} Optimize scale $s_{\text{MRAM}}$ by solving:
\[
s_{\text{MRAM}}^* = \arg\min_{s}\;\|W_{\text{out}}-Q(W_{\text{out}};s)\|^2
\]
then quantize
\[
W^*_{\text{out}}=Q(W_{\text{out}};s_{\text{MRAM}}^*).
\]

\STATE \textbf{Step 4: Merge Results} Reconstruct  
$\tilde{W}=\text{scatter}(W^*_{\text{in}},W^*_{\text{out}})$.

\RETURN $\tilde{W}$
\end{algorithmic}
\end{algorithm}

\begin{align}
\mathcal{L}(s)
&= \mathbb{E}_{e}\!\left[\|W_{\text{in}} - (Q(W_{\text{in}};s)+e)\|_2^2\right] \\[2pt]
&\approx \|W_{\text{in}} - Q(W_{\text{in}};s)\|_2^2 + \sum_i \mathbb{E}[e_i^2] 
\quad\!\! (\mathbb{E}[e_i]\approx0) \\[2pt]
&\approx \|W_{\text{in}} - Q(W_{\text{in}};s)\|_2^2
   + |W_{\text{in}}|\,(p_{-}+p_{+})\,\Delta(s)^2 .
\end{align}

\noindent where $|W_{\text{in}}|$ denotes the total number of inlier weights being stored in ReRAM. Hence, $\mathcal{L}(s)$ depends on $s$ both through the quantized weights and through the noise amplitude $\Delta(s)$. In practice, we evaluate this one-dimensional objective over a grid of candidate scales and select the minimizer $s_{\text{ReRAM}}^*$, balancing standard quantization distortion against sensitivity to ReRAM-induced noise.

Outliers, on the other hand, follow Step~3, where their scale $s_{\text{MRAM}}^*$ is optimized using a standard MSE objective. By explicitly incorporating BER-driven perturbation models only into the quantization of inliers, our design achieves compression efficiency while preserving accuracy under hardware-induced noise.
Unlike prior ReRAM-based noise-aware training schemes that incorporate device non-idealities during quantization-aware training~\cite{cherupally2022improving,yang2021multi}, our approach keeps the pretrained model fixed and applies noise awareness solely in the post-training quantization stage, yielding a training-free method robust to device variability.


\subsection{Orthogonality to Existing PTQ}
Finally, we emphasize that QMC is \emph{orthogonal} to existing PTQ frameworks. This complementarity nature enables QMC to be combined with methods to deliver state-of-the-art accuracy while also enhancing hardware efficiency. Thus, QMC serves as a practical building 
block for the next generation of quantization pipelines.

\section{Results}

\begin{table}[tp]
    \centering
    \renewcommand{\arraystretch}{1}  
    \setlength{\tabcolsep}{6pt} 
    \caption{MRAM, ReRAM, and LPDDR5 characteristics Table}
    \resizebox{\linewidth}{!}{  
\begin{tabular}{lcccc}
    \toprule
    & \textbf{Operation} 
    & \textbf{MRAM}~\cite{zhang2022cache, dieny2020opportunities} 
    & \textbf{MLC ReRAM}~\cite{jain201913, fan2024efficient} 
    & \textbf{LPDDR5}~\cite{SamsungLPDDR5} \\
    \midrule
    \textbf{Latency (ns)} & Read  & 3.5  & $<$5 & 1.7 \\
    \midrule
    \textbf{Bandwidth (GiB/s)} & Read  & 36.57 (per channel)  & 1.8 (per 256$\times$256 array) & 186.26 \\                                
    \midrule
    \textbf{Energy (pJ/bit)} & Read  & 1  & 1.56 (3-bit mode) & 3.5 \\
    \midrule
    \textbf{Density (Mb/mm\textsuperscript{2})} & -- & 66  & 30.1 (3-bit mode) & 209.9 \\
    \midrule
    \textbf{Process Node (nm)} & -- & 5  & 22 & D1y \\
    \bottomrule
\end{tabular}   
    } 
    \label{tab:memory_comparison}
    \vspace{-0.3cm}
\end{table}

\subsection{Experimental Setup}

\begin{table*}[t]
\centering
\scriptsize
\fontsize{7}{7}\selectfont
\setlength{\tabcolsep}{7pt}
\caption{Comparison of FP16, RTN INT4, MXINT4, and our method. 
WikiText perplexity ($\downarrow$) lower is better and reasoning benchmarks ($\uparrow$) higher is better). 
Compression ratios are relative to FP16.}

\renewcommand{\arraystretch}{0.6}
\setlength{\tabcolsep}{10pt}   
\begin{tabular}{llcccccc}
\hline
\textbf{Model} & \textbf{Config} & \textbf{Wikitext PPL $\downarrow$} & \textbf{Hella $\uparrow$} & \textbf{BoolQ $\uparrow$} & \textbf{ARC-e $\uparrow$} & \textbf{ARC-c $\uparrow$} & \textbf{Compression Ratio} \\
\hline
\multirow{6}{*}{Hymba-Instruct-1.5B}
 & FP16 & 11.87 & 71.10 & 82.14 & 76.14 & 48.89 & 1× \\
 & RTN INT4~\cite{nagel2020up} & 49.10 & 55.65 & 68.23 & 60.52 & 36.86 & 4× \\
 & MXINT4~\cite{sharify2024post} & 21.63 & 63.33 & 76.73 & 66.25 & 41.13 & 4× \\
 & QMC (3bits-MLC) & 13.27 & 69.35 & 80.61 & 75.42 & 47.35 & 4.44× \\
 & QMC (2bits-MLC) & 12.54 & 70.06 & 82.07 & 75.96 & 48.63 & 4.44× \\
\hline
\multirow{7}{*}{LLaMA-3.2-3B}
 & FP16 & 9.27 & 73.65 & 72.84 & 71.68 & 45.90 & 1× \\
 & RTN INT4~\cite{nagel2020up} & 20.93 & 61.24 & 65.20 & 61.49 & 35.84 & 4× \\
 & MXINT4~\cite{sharify2024post} & 18.90 & 60.99 & 69.57 & 64.98 & 36.01 & 4× \\
 & QMC (3bits-MLC) & 12.77 & 68.40 & 70.49 & 66.41 & 40.36 & 4.44× \\
 & QMC (2bits-MLC) & 10.81 & 71.77 & 72.05 & 72.31 & 45.05 & 4.44× \\
\hline
\multirow{6}{*}{Phi-1.5B}
 & FP16 & 29.16 & 62.68 & 74.56 & 73.32 & 48.12 & 1× \\
 & RTN INT4~\cite{nagel2020up} & 36.44 & 60.16 & 73.79 & 70.83 & 46.25 & 4× \\
 & MXINT4~\cite{sharify2024post} & 34.13 & 60.06 & 70.95 & 71.13 & 44.80 & 4× \\
 & QMC (3bits-MLC) & 32.18 & 60.46 & 59.60 & 71.13 & 45.56 & 4.44× \\
 & QMC (2bits-MLC) & 30.47 & 61.79 & 69.88 & 72.14 & 45.99 & 4.44× \\
\hline
\multirow{6}{*}{Qwen2.5-1.5B-Instruct}
 & FP16 & 12.20 & 68.25 & 78.20 & 75.80 & 46.93 & 1× \\
 & RTN INT4~\cite{nagel2020up} & 29.14 & 59.42 & 60.28 & 63.97 & 40.10 & 4× \\
 & MXINT4~\cite{sharify2024post} & 20.68 & 61.39 & 69.88 & 69.65 & 43.26 & 4× \\
 & QMC (3bits-MLC) & 17.79 & 63.49 & 64.98 & 71.38 & 45.82 & 4.44× \\
 & QMC (2bits-MLC) & 14.61 & 66.42 & 68.84 & 72.85 & 45.56 & 4.44× \\
\hline
\end{tabular}
\label{tab:main-results}
\end{table*}

\textbf{Quantization Setup:} We conduct all experiments using uniform per-channel quantization, the default mode supported by most commercial edge platforms~\cite{nagel2021white,nvidia2023tensorrt}. Inlier weights are mapped to either 3- or 2-bits MLC ReRAM using the proposed noise-aware quantizer, while outliers are quantized using 5 bits and stored in MRAM. We vary the outlier ratio $\rho$ to study its effect on accuracy and compression. 


\begin{table}[t]
\centering
\scriptsize
\caption{Quantization method comparison.}
\renewcommand{\arraystretch}{1}
\fontsize{7}{7}\selectfont
\setlength{\tabcolsep}{4pt}
\begin{tabular}{llcccccc}
\hline
\textbf{Model} & \textbf{Method} & \textbf{PPL$\downarrow$} & \textbf{Hella$\uparrow$} & \textbf{BoolQ$\uparrow$} & \textbf{ARC-e$\uparrow$} & \textbf{ARC-c$\uparrow$} \\
\hline
\multirow{3}{*}{\tiny LLaMA-3.2B}
 & AWQ~\cite{lin2024awq} & 12.67 & 69.91 & 74.29 & 67.72 & 44.80 \\
 & GPTQ~\cite{frantar2022gptq} & 10.57 & 72.17 & 70.73 & 69.74 & 44.28 \\
 & QMC (no noise) & 10.43 & 72.62 & 71.28 & 72.85 & 46.16 \\
\hline
\multirow{3}{*}{\tiny Qwen2.5-1.5B-Instruct}
 & AWQ~\cite{lin2024awq} & 13.25 & 66.40 & 74.25 & 72.43 & 44.37 \\
 & GPTQ~\cite{frantar2022gptq} & 13.48 & 65.84 & 75.44 & 75.88 & 45.14 \\
 & QMC (no noise) & 13.89 & 66.83 & 75.29 & 73.95 & 45.22 \\
\hline
\end{tabular}
\label{tab:algo-compare}
\end{table}
\textbf{NVM Noise model and Memory system simulators:} To establish the MLC ReRAM error model, we modified the DL-RSim framework~\cite{lin2018dl} to model the noise characteristics shown in Figure~\ref{fig:error_model}. The validity of this proposed model was verified against measured error distributions from MLC ReRAM devices~\cite{fan2024efficient}.
For performance and energy evaluations, we extended NVMain~\cite{poremba2015nvmain} to support memory simulations for LLM workloads. The modified simulator tracks overall latency and energy consumption during model weight transactions, enabling a comprehensive analysis of design trade-offs in QMC. We use state-of-the-art MRAM and ReRAM technologies using the specifications summarized in Table~\ref{tab:memory_comparison}. 


\textbf{Baselines:} We evaluate our approach against four categories of baselines:

\begin{itemize}
    \item \textit{Algorithm-Level Baselines:} We benchmark against state-of-the-art PTQ methods such as AWQ~\cite{lin2024awq} and GPTQ~\cite{frantar2022gptq}.

    \item \textit{System-Level Baselines:} We include FP16, Rounding-to-Nearest (RTN) INT4~\cite{nagel2021white}, and MXINT4~\cite{sharify2024post}, formats commonly used in GPU or accelerator with DRAM-based memory hierarchies.

    \item \textit{Hardware Platform Baseline:} As the conventional edge device, we adopt the Jetson AGX Orin~\cite{orin_agx_module_ds}, which uses LPDDR5~\cite{SamsungLPDDR5} as main memory and reflects a widely deployed configuration in edge AI systems.

    \item \textit{Hardware/Algorithm Co-Design Baseline:} We benchmark against eMEMs~\cite{mukherjee2021case}, the state-of-the-art architecture leveraging emerging memory technologies for storing weights.
\end{itemize}


\begin{figure}
  \centering
  \includegraphics[width=0.9\columnwidth]{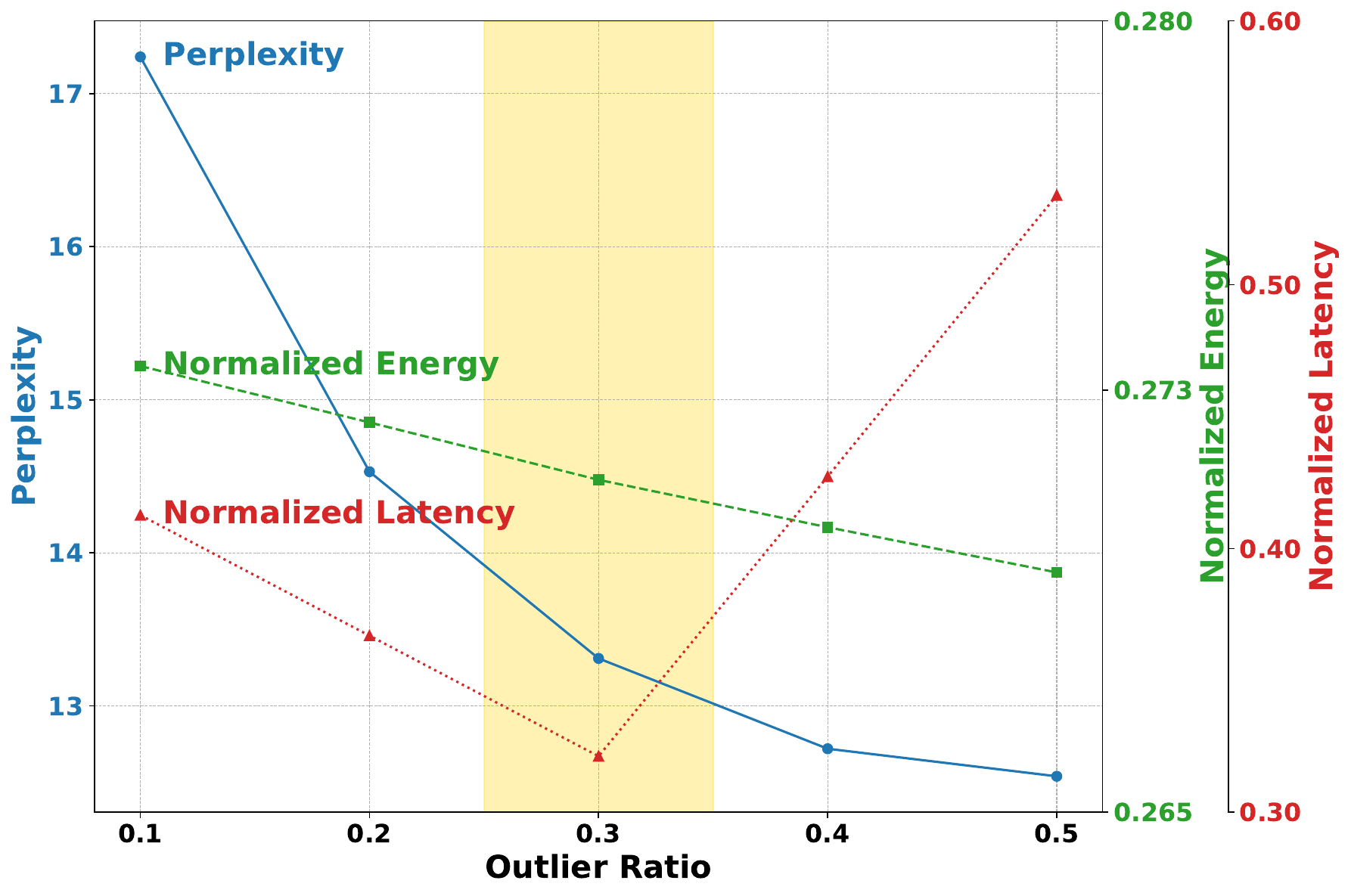}
    \caption{\textbf{Relationship between outlier ratio and perplexity, and the corresponding normalized energy/latency.}}
  \vspace{-1\baselineskip} 
  \label{fig:optimized_point}
\end{figure}


\textbf{Models and Datasets:} We evaluate our method on WikiText~\cite{merity2018scalable} dataset. 
For reasoning, we use Hellaswag (Hella)~\cite{zellers2019hellaswag}, BoolQ~\cite{clark2019boolq}, and ARC-Easy (ARC-e) and Challenge (ARC-c)~\cite{clark2018think} subsets on popular Small Language Models (SLMs) Hymba-Instruct-1.5B~\cite{dong2024hymba}, Qwen2.5-1.5B-Instruct~\cite{team2024qwen2}, LLaMA-3.2-3B~\cite{touvron2023llama}, and Phi-1.5B~\cite{javaheripi2023phi}.

\textbf{Evaluation Metrics:} We evaluate our method using three categories of metrics. For language modeling quality, we report perplexity (PPL) on WikiText~\cite{merity2018scalable}, where lower PPL indicates better performance. For reasoning tasks, we measure accuracy. Finally, to capture system-level behavior, we report end-to-end energy and latency under both 3-bit and 2-bit MLC ReRAM modes, reflecting real-world deployment characteristics.

\vspace{-0.05in}


\begin{figure*}[t]
  \centering
  \includegraphics[width=0.95\textwidth]{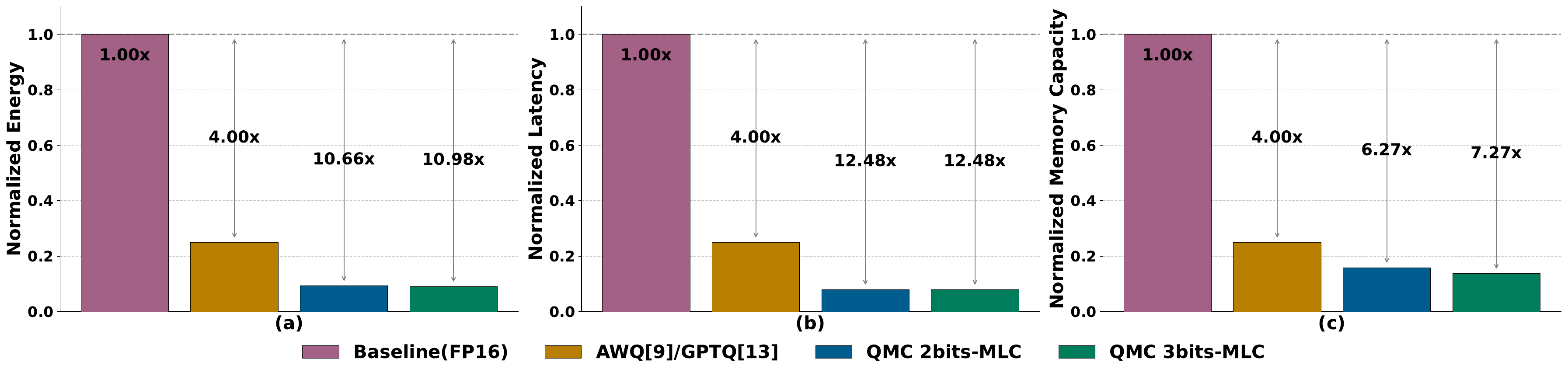}
    \caption{\textbf{Quantization Impact on System Performance.} Energy (left), latency (center), and memory capacity (right) for Hymba~1.5B~\cite{dong2024hymba} on WikiText~\cite{merity2018scalable}. QMC (2-bit/3-bit MLC) achieves 6.27$\times$--12.48$\times$ gains over FP16 and outperforms AWQ/GPTQ~\cite{lin2024awq,frantar2022gptq}. FP16 and existing PTQ baselines use LPDDR5-based systems. While QMC uses the proposed heterogeneous NVM hierarchy.
}

  \vspace{-1\baselineskip} 
  \label{fig:AllComp}
\end{figure*}

\vspace{-0.05in}
\subsection{Experimental Results}
\label{subsec:results}

\subsubsection{Impact of Outlier Ratio}
\label{subsec:outlier_ratio_ablation}
We define the outlier ratio $\rho$ as the fraction of large-magnitude weights designated as outliers and retained at higher precision ($Q_{\text{high}}$), while the remaining are treated as inliers and aggressively quantized ($Q_{\text{low}}$). This parameter is central to QMC, as it governs the balance between accuracy preservation and system efficiency. Figure~\ref{fig:optimized_point} illustrates the effect of varying $\rho$ on both model quality and runtime efficiency. Increasing $\rho$ consistently improves perplexity (from 17.24 at $\rho=0.1$ to 12.54 at $\rho=0.5$), while latency decreases until $\rho=0.3$ but rises at $\rho=0.4$ as MRAM access becomes the bottleneck. Overall, a moderate ratio ($\rho=0.3$) yields a balanced trade-off, achieving near-optimal accuracy while minimizing efficiency loss. Although we present results on Wikitext, similar patterns are observed on other benchmarks.

\subsubsection{Algorithm-Only Evaluation Analysis}
\label{subsec:algo_only}


\paragraph{\textbf{Algorithmic results:}}
Across the two evaluation settings, our method consistently outperforms existing PTQ baselines. 
In the \emph{algorithm-only} view (Table~\ref{tab:algo-compare}), QMC achieves lower perplexity 
and higher downstream accuracy than AWQ~\cite{lin2024awq} and GPTQ~\cite{frantar2022gptq}, 
without requiring calibration data. For example, on LLaMA-3.2B we reduce perplexity below 
both AWQ and GPTQ while improving accuracy across multiple tasks. This gain arises from targeting 
important weights and retaining outliers in higher precision while quantizing inliers more 
aggressively, thereby reducing quantization noise more effectively than data-dependent approaches. 
In the \emph{system-level} view (Table~\ref{tab:main-results}), QMC substantially narrows the gap 
to FP16 compared to RTN INT4 and MXINT4~\cite{sharify2024post}. For instance, on Hymba-Instruct, 
RTN INT4 degrades performance sharply, while QMC remains close to the FP16 baseline with greater 
compression ratio (4.44x). Similar trends hold across LLaMA-3.2-3B, Phi-1.5B, and Qwen2.5-1.5B-Instruct. Overall, 
QMC improves accuracy–compression trade-offs while remaining framework-agnostic and deployable, 
in contrast to AWQ and GPTQ, which lack software support for newer compact models such as Hymba and Phi.

Next, we now evaluate \emph{system-level} energy, latency, and capacity under ReRAM–MRAM assumptions.

\subsubsection{System-Level Performance Analysis}
\label{subsec:sys_only}
\paragraph{\textbf{Energy Efficiency Analysis:}}
Figure~\ref{fig:AllComp}(a) presents energy consumption on Hymba-Instruct-1.5B model and WikiText benchmark. QMC achieves $10.98\times$ energy reduction with 3-bit MLC compared to the FP16 baseline and outperforms existing quantization methods AWQ~\cite{lin2024awq} and GPTQ~\cite{frantar2022gptq} (Figure~\ref{fig:AllComp}). Three factors drive these improvements: (1) \textbf{data volume reduction} by $4.44\times$ through aggressive quantization, (2) \textbf{elimination of off-chip transfers} with 87\% reduction in DRAM access, and (3) \textbf{lower read energy} of MRAM/ReRAM compared to DRAM (Table~\ref{tab:memory_comparison}). The combination of embedded storage for critical outlier weights and aggressive quantization of inlier weights in high-density ReRAM enables this significant energy reduction.

Figure~\ref{fig:optimized_point} reveals minimal energy variation across outlier ratios due to an inherent trade-off: while higher ratios reduce off-chip transfers through increased embedded memory usage, the transition from 3-bit inlier to 5-bit outlier quantization increases total data volume and requires additional processing to handle mixed-precision formats, consequently raising MRAM read energy and on-chip transfer costs. This counterbalances the benefits of reduced off-chip access, resulting in relatively flat energy consumption.

\paragraph{\textbf{Performance Efficiency Analysis:}}
Figures~\ref{fig:AllComp}(b) show QMC achieves $12.48\times$ latency reduction versus FP16 and $3.12\times$ versus AWQ/GPTQ~\cite{lin2024awq,frantar2022gptq}. Two factors enable these gains: (1) \textbf{$4.44\times$ data reduction} through Outlier-Aware Quantization, directly decreasing memory access time, and (2) \textbf{optimized memory balancing} between MRAM and ReRAM within power constraints. Systematic exploration identified the optimal sweet spot at 0.3 outlier ratio where bandwidth utilization is balanced. Deviations disrupt this balance, creating bottlenecks that explain the U-shaped latency curve in Figures~\ref{fig:optimized_point} and, validating our co-design approach.
\paragraph{\textbf{Memory Capacity and Area Efficiency Analysis:}}
Non-volatile memories enable storage-memory consolidation, eliminating Flash-DRAM separation. QMC achieves $7.27\times$ memory cell reduction versus FP16 through aggressive quantization and 3-bit MLC ReRAM, with $14.54\times$ reduction versus traditional LPDDR5+Flash architecture. Area analysis for the Hymba-Instruct-1.5B model~\cite{dong2024hymba} shows eliminating Flash/DRAM saves 112.04mm² while ReRAM/MRAM requires 133.66mm², yielding a net increase of 21.62mm². 

\begin{table}[t]
\centering
\caption{Performance Comparisons of Co-Design Methods. Lower is better }
\label{tab:comparison_results}
\resizebox{\columnwidth}{!}{%
\begin{tabular}{lcccc}
\toprule
\textbf{Configuration} & \textbf{\begin{tabular}[c]{@{}c@{}}Norm.\\ Energy\end{tabular}} & \textbf{\begin{tabular}[c]{@{}c@{}}Norm.\\ Latency\end{tabular}} & \textbf{\begin{tabular}[c]{@{}c@{}}Norm.\\ Capacity\end{tabular}} & \textbf{PPL}$\downarrow$ \\ \midrule
eMEMs with MRAM~\cite{mukherjee2021case} & $0.96\times$ & $1.90\times$ & $1.82\times$ & 20.93 \\
eMEMs with MLC ReRAM~\cite{mukherjee2021case} & $1.35\times$ & $1.90\times$ & $0.61\times$ & 24.71 \\
\textbf{QMC} & $\mathbf{1}\times$ & $\mathbf{1}\times$ & $\mathbf{1}\times$ & \textbf{12.77} \\ \bottomrule
\end{tabular}%
}
\end{table}
\paragraph{\textbf{Co-Design Method Comparison:}}
Table \ref{tab:comparison_results} presents the performance and accuracy comparison with existing emerging memory co-design methods, eMEMs~\cite{mukherjee2021case}. For energy, eMEMs utilizing MRAM records a lower energy consumption than QMC, which is attributed to MRAM's intrinsically low read energy as detailed in Table \ref{tab:memory_comparison}. In contrast, eMEMs with MLC ReRAM exhibits the highest energy consumption as its higher read energy. Regarding latency, QMC enables it to achieve $1.9 \times$ lower latency compared to eMEMs thanks to aggressive quantization and hybrid memory architecture which reduces the data transfer time. eMEMs relies solely on off-chip emerging memory and lower quantization ratio, resulting in worse performance than QMC. For capacity and PPL, eMEMs with MLC ReRAM offers the best memory capacity as it stores three bits of data per memory cell. However, its quantization method is highly susceptible to noise, leading to the worst PPL performance. Overall, QMC demonstrates the best comprehensive performance when considering both performance metrics and accuracy.
\vspace{-0.1in}

\paragraph{\textbf{System Overhead Analysis:}}
Our QMC framework introduces several implementation overheads that must be weighed against its efficiency gains. The heterogeneous memory design incurs a 21.62\,mm$^2$ area overhead due to the lower density of MRAM and ReRAM relative to DRAM and Flash (Table~\ref{tab:memory_comparison}). Clock-domain crossing between the two memory tiers requires dual-clock FIFO synchronizers, adding 2--4 cycles of latency and 1--2\,mW of power~\cite{apperson2007scalable}. When using 2-bit MLC ReRAM for improved noise tolerance, additional cost arises from bit packing/unpacking due to the mismatch between 3-bit weight quantization and 2-bit cell storage. Despite these overheads, the resulting 10.98$\times$ energy savings and 12.48$\times$ performance improvement outweigh these overheads for edge AI scenarios where efficiency is prioritized over area.

\section{Conclusion}

We introduced QMC, a novel outlier-aware quantization framework co-designed with heterogeneous non-volatile memory to enable efficient SLM deployment on edge platforms. Across language modeling and reasoning benchmarks, QMC delivers $6.27\times$–$7.27\times$ memory reduction, $7.62\times$ lower data movement, and up to $10.98\times$ energy and $12.48\times$ latency improvements over FP16, while maintaining accuracy on par with state-of-the-art quantization methods. QMC also outperforms INT4 and MXINT4 baselines,even under realistic ReRAM noise without requiring retraining. By storing outliers in high-precision MRAM and aggressively quantizing inliers into MLC ReRAM, QMC achieves a favorable accuracy–efficiency trade-off that makes SLMs practical for real-time, privacy-sensitive edge inference. Overall, QMC offers a scalable, deployment-ready co-design that tightly integrates quantization and memory architecture for next-generation generative AI systems.

\begin{acks}
This work has been funded in part by NSF, with award numbers \#1826967, \#1911095, \#2003279, \#2052809, \#2100237, \#2112167, \#2112665, and in part by PRISM and CoCoSys, centers in JUMP 2.0, an SRC program sponsored by DARPA. We also thank Kiseok Suh from Samsung for his valuable technical support and feedback that helped improve this work.
\end{acks}

\bibliographystyle{IEEEtran}

\end{document}